\def\year{2022}\relax
\documentclass[letterpaper]{article} 
\usepackage{aaai22}  
\usepackage{times}  
\usepackage{helvet}  
\usepackage{courier}  
\usepackage[hyphens]{url}  
\usepackage{graphicx} 
\usepackage{natbib}  
\usepackage{caption} 
\DeclareCaptionStyle{ruled}{labelfont=normalfont,labelsep=colon,strut=off} 
\usepackage{algorithm}
\usepackage{algorithmic}

\usepackage{amsmath}
\usepackage{amssymb}
\usepackage{makecell}
\usepackage{multirow}
\usepackage{multicol}
\usepackage{booktabs}
\usepackage{pifont}

\usepackage{arydshln}
\usepackage[dvipsnames]{xcolor}
\usepackage{newfloat}
\usepackage{listings}
\floatstyle{ruled}
\newfloat{listing}{tb}{lst}{}
\floatname{listing}{Listing}
\usepackage[normalem]{ulem}

\title{SyncTalkFace: Talking Face Generation with \\ Precise Lip-Syncing via Audio-Lip Memory}
\author{
    Se Jin Park, Minsu Kim, Joanna Hong, Jeongsoo Choi, Yong Man Ro\thanks{Corresponding author.}
}
\affiliations{
Image and Video Systems Lab, KAIST, South Korea\\
\{jinny960812, ms.k, joanna2587, jeongsoo.choi, ymro\}@kaist.ac.kr
    
}

\usepackage{bibentry}

\begin{document}

\maketitle

\begin{abstract}
The challenge of talking face generation from speech lies in aligning two different modal information, audio and video, such that the mouth region corresponds to input audio. Previous methods either exploit audio-visual representation learning or leverage intermediate structural information such as landmarks and 3D models. However, they struggle to synthesize fine details of the lips varying at the phoneme level as they do not sufficiently provide visual information of the lips at the video synthesis step. To overcome this limitation, our work proposes Audio-Lip Memory that brings in visual information of the mouth region corresponding to input audio and enforces fine-grained audio-visual coherence. It stores lip motion features from sequential ground truth images in the value memory and aligns them with corresponding audio features so that they can be retrieved using audio input at inference time. Therefore, using the retrieved lip motion features as visual hints, it can easily correlate audio with visual dynamics in the synthesis step. By analyzing the memory, we demonstrate that unique lip features are stored in each memory slot at the phoneme level, capturing subtle lip motion based on memory addressing. In addition, we introduce visual-visual synchronization loss which can enhance lip-syncing performance when used along with audio-visual synchronization loss in our model. Extensive experiments are performed to verify that our method generates high-quality video with mouth shapes that best align with the input audio, outperforming previous state-of-the-art methods. \end{abstract}

\section{Introduction}
Talking face generation from speech, also referred to as lip-syncing, is synthesizing a video of a target identity such that the mouth region is consistent with arbitrary audio input. It has many applications such as audio-driven photorealistic avatars that can be employed in online classes or games, dubbing films in another language, and communication aids for the hearing-impaired who can lip-read. As the talking face generation carries various practical usage, it has received great interest for research. 

The main challenge of talking face generation from speech is aligning audio and visual information so that the generated facial sequence is coherent with the input audio. Previous methods based on encoder-decoder structure have worked on improving audio and visual representations. \cite{zhou2019talking, zhou2021pose} disentangled visual input into identity and speech content space using metric learning and enhanced audio feature by embedding into a shared latent space between visual feature. \cite{mittal2020animating} disentangled audio representation into phonetic content, emotional tone, and other factors. They have explored feature disentanglement to remove irrelevant factors in lip-syncing. However, the disentangled audio representation does not explicitly contain visual information of the mouth which can help the decoder to map visual dynamics from audio. 

Recent advances utilize intermediate structural representations such as facial landmarks and 3D models to better capture facial dynamics. \cite{chen2019hierarchical, das2020speech, zhou2020makelttalk} mapped lip landmarks from audio and composited into the mouth region of a target person.  \cite{song2020everybody, thies2020neural} learned speaker independent features in a 3D face model and rendered a talking face video of a target person with fine-tuning. However, they commonly lack sophistication in lip-syncing. This is because the facial landmarks are too sparse to provide accurate lip synchronization, and the 3D models cannot capture fine details in the mouth region including teeth \cite{zhang2021flow, wang2021one}. Also, they bear the limitation of having to acquire the intermediate representation separately from the generation network.
 
Distinct from the previous works, we introduce Audio-Lip Memory that explicitly provides visual information of the mouth region and enables more precise lip synchronization with input audio. The memory learns to align audio with corresponding lip features from sequential ground truth images during training, so that it outputs the audio-aligned lip features, when queried with audio at inference time. The recalled lip features are fused with audio features and injected into the decoder for synthesizing the talking face video. As the decoder can leverage the explicit visual hints of the mouth, it can better map audio to video both temporal- and pixel-wise. Moreover, the Audio-Lip Memory stores the representative lip features at the phoneme level and retrieves various combinations of the lip features through memory addressing, enabling sophisticated and diverse lip movements. We additionally impose visual-visual synchronization along with audio-visual synchronization for strong lip-syncing. Hence, the proposed model achieves high-quality video generation with fine-grained audio-visual coherence.

Our contributions are as follows: (1) We propose Audio-Lip Memory that maps audio to lip-movement intermediate representations that bridge audio with lip sync video generation. It explicitly provides visual hints of the lip movement to the decoder and enhances the sophistication of lip motion corresponding to the audio. (2) We ensure strong lip synchronization by utilizing visual-visual synchronization between ground truth face sequence and generated face sequence along with audio-visual synchronization between input audio and generated faces. (3) By analyzing learned representations inside the memory, we confirm that the representations are stored at the phoneme level in each memory slot and different combinations of addressing slots yield variational mouth shapes. Thus, direct manipulation of lip movement using memory address is possible. (4) We achieve state-of-the-art performance on LRW and LRS2 dataset in terms of visual quality and lip-sync quality.

\begin{figure*}[t!]
	\begin{minipage}[b]{1.0\linewidth}
		\centering
		\centerline{\includegraphics[width=15cm]{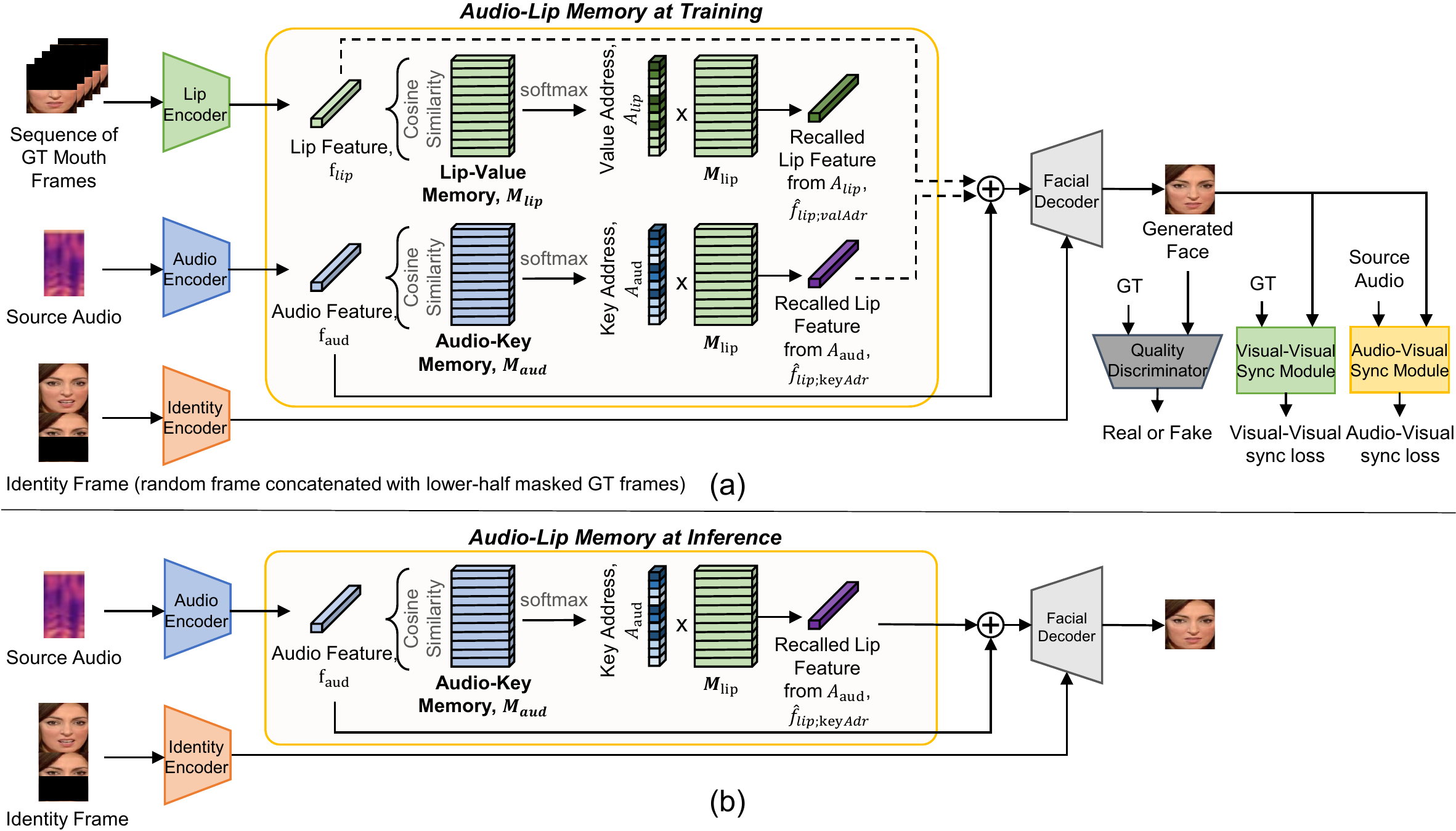}}
	\end{minipage}
	\caption{Overview of our proposed model. During training, Audio-Lip Memory learns to store lip feature $f_{lip}$ in the lip-value memory and to align key address $A_{aud}$ with value address $A_{lip}$ as depicted in (a). During inference, the model utilizes recalled lip feature from key address, $\hat {f}_{lip;keyAdr}$, obtained from audio input as a query as shown in (b).}
	\vspace{-0.2cm}
	\label{fig:1}
\end{figure*}

\section{Related Work}
\subsubsection{Talking Face Generation}
Existing works on talking face generation can be broadly categorized into reconstruction based methods and intermediate representation based methods. Reconstruction based methods \cite{chen2018lip, song2018talking, jamaludin2019you, kr2019towards, vougioukas2020realistic} follow the encoder-decoder structure where identity features and speech features are extracted and fused together as an input to a decoder to synthesize talking face videos in an end-to-end manner. \cite{prajwal2020lip} took a face video as a visual input and used lower-half masked of the input video as a pose prior. It employed a pre-trained lip-sync discriminator and highlighted the importance of an accurate lip-sync discriminator that can feedback lip-sync quality to the network. \cite{zhou2019talking} disentangled speech related features and identity related features from video input through associative-and-adversarial training. In \cite{zhou2021pose}, the author further disentangled visual input into identity space, pose space, and speech content space, allowing free pose-control. Although many works have explored improving visual representation by disentangling different factors in visual input, not much work has sought into improving audio representation. \cite{mittal2020animating} attempted to improve performance from the perspective of audio representation learning. They disentangled phonetic content, emotional tone, and the rest of the other factors from audio using Variational Autoencoder with KL divergence and negative log likelihood with margin ranking loss. Instead of decoupling speech related features from the audio, our work explicitly filters out lip motion related features from the input audio. As we directly map audio to lip features before injecting the audio features to the generator, we can impose lip synchronization earlier in the generation step. 

Intermediate representation based methods consist of two cascaded modules where intermediate representations such as landmarks and 3D models are leveraged to generate video from input audio. \cite{chen2019hierarchical, das2020speech} estimated facial landmarks from input audio and then generated video conditioned on the generated landmarks and a reference image. \cite{zhou2020makelttalk, das2020speech} separately considered speech content related landmarks and speaker identity related landmarks for the generation of unseen subjects. 3D model based methods commonly extract expression, geometry, and pose parameters to reconstruct 3D facial mesh (3DMM) from which a face video is generated \cite{song2020everybody, yu2020multimodal}.  \cite{thies2020neural} used a pretrained audio-expression network to model an expression basis in the 3D face model. \cite{zhang2021flow} proposed a style-specific generator that produces facial animation parameters that are combined with facial shape parameters to create 3D mesh points. Such intermediate representations provide structural information of facial dynamics that has limitations in containing fine details of the mouth. Moreover, acquiring the landmarks and 3D models is laborious and time-consuming. We try to overcome these limitations by leveraging recalled lip features from memory. The memory stores lip features in value memory slots at the phoneme level during training so that information about lip motion corresponding to input audio can be obtained at inference. Also, as various combinations of the lip features in each slot are possible through memory addressing, more diverse and subtle lip movements can be portrayed. 

\subsubsection{Audio-Visual Alignment} Audio-visual alignment aims to find the correlation space between audio and video, and find temporal coherence between the two modality data. In the context of talking face generation task, \cite{prajwal2020lip} directly employed a pretrained embedding module \cite{chung2016out} as a lip-sync discriminator, and \cite{inproceedings} presented asymmetric Mutual Information Estimator. They all relied on the audio-visual embedding module placed at the end of the generator network to give feedback to the whole network on the coherence between input audio and generated video. More recent works on cross-modal learning apply multi-way matching loss that considers intra-class pairs as well as inter-class pairs, and have shown its effectiveness \cite{chung2019perfect, nagrani2020disentangled, gao2021visualvoice}. Inspired by the intra-class loss, our work additionally exploits visual-visual sync loss. As input audio and ground truth video are in sync, we can expect the complementary effect of audio-visual alignment by aligning visual lip features from generated face sequence and ground truth face sequence.

\subsubsection{Memory Network}
Memory Network \cite{weston2014memory} provides a long-term memory component that can be read from and written to with inference capability. \cite{miller2016key} introduced key-value memory structure where key memory is used to address memories with respect to a query and corresponding value is obtained from value memory using the address. Since the scheme can remember selected information, it is effective for augmenting features \cite{kaiser2017learning, lee2018memory, cai2018memory, zhu2019dm, pei2019memory, lee2021video, kim2021robust, kim2021multi, kim2021m}. \cite{yi2020audio} incorporated memory to talking face generation to refine roughly rendered frames into realistic frames. It stores spatial features and identity features as key-value pairs and retrieves the best-matching identity feature using the spatial feature as a query. Unlike previous works that use memory only to remember critical information, we employ the key-value memory to align and store two different modality features. We map audio to lip features through key-value memory addressing so that the lip features not available at inference can be utilized with audio input as a query. The recalled lip features from the value memory is used as an intermediate representation to bridge between audio and video.

\section{Methods}
We propose Audio-Lip Memory that explicitly maps audio to lip features. Our whole pipeline is depicted in Fig.\ref{fig:1}. We take a frame of target identity, 0.2 seconds of source audio, and upper half masked face sequence (5 frames) corresponding to the source audio as input. We aim to lip-sync the input video of the target identity such that the mouth region is consistent with the input audio, altering only the mouth while preserving all the other elements (i.e., pose, identity, and etc). We align and store encoded audio features and encoded lip features in the Audio-Lip Memory so that lip features can be obtained when queried with audio features. The recalled lip features from the memory are fused with audio features and injected into the decoder network for video synthesis. In addition, strong lip synchronization is imposed with audio-visual synchronization loss and visual-visual synchronization loss. We explain the details of the Audio-Lip memory in Sec. 3.1 and video synthesis in Sec. 3.2. 

\subsection{Audio-Lip Memory} 
Audio-Lip Memory maps audio features to lip motion related features. We firstly encode a spectrogram of source audio (0.2 seconds) into audio feature $f_{aud} \in \mathbb{R}^{C}$ using an audio encoder, and corresponding 5 sequential frames with upper half masked to lip feature $f_{lip} \in \mathbb{R}^{C}$ using a lip encoder. Audio-Lip Memory is composed of an audio-key memory $ \mathbf M_{aud} \in \mathbb{R}^{S \times C}$ and a lip-value memory $\mathbf M_{lip} \in \mathbb{R}^{S \times C}$, where $S$ denotes slot size and $C$ channel. Note that we universally set $C$ to 512. The memory learns to store representative lip features in the lip-value memory through reconstruction loss between recalled lip features from the key address and lip features extracted from the lip encoder. It simultaneously learns to align lip features with audio features through key-value address alignment loss so that the corresponding lip feature can be retrieved using an audio feature as a query. 

\subsubsection{Storing lip features in lip-value memory}
The lip-value memory $\mathbf M_{lip}=\{m_{ lip}^{i}\}_{i=1}^{S}$ where $m_{lip}^{i} \in \mathbb{R}^{C}$ is a unique lip feature in the $i$-th slot.
When a lip feature of 5 sequential mouth frames from the lip encoder is given as a query, distance between the lip feature and each of the slots is computed using cosine similarity: 
\begin{equation}
d_{lip}^{i}=\frac{m_{lip}^{i} \cdot f_{lip}}{\|m_{lip}^{i}\|_{2} \cdot\|f_{lip}\|_{2}} . 
\end{equation}
Then, we take softmax of the similarity distance computed on individual slots as follows:
\begin{equation}
\alpha_{lip}^{i}=\frac{\exp (\kappa \cdot d_{lip}^{i})}{\sum_{j=1}^{S} \exp (\kappa \cdot d_{lip}^{j})} ,
\end{equation}
where $\kappa$ is a scaling term, and $\alpha_{lip}^{i}$ is an attention weight on the $i$-th slot of the lip-value memory with respect to the lip feature. By computing the attention weights for all slots, we get a value address $A_{lip}=\{\alpha_{lip}^{1}, \alpha_{lip}^{2}, \ldots \alpha_{lip}^{S}\} \in \mathbb{R}^{S} $. 
It is used to locate relevant slots in the lip-value memory associated with the lip feature. 
Finally, we can retrieve the lip feature associated with the query by taking dot product between the value address and the lip-value memory:  
\begin{equation}
\hat{f}_{{lip;valAdr}}=A_{ {lip}} \cdot \mathbf{M}_{ {lip}} .
\end{equation}
We denote $\hat{f}_{ {lip;valAdr}} \in \mathbb{R}^{C}$ as recalled lip feature from the value address. By taking the weighted sum of the different lip features stored in individual slots, we can utilize various combinations of the lip features and generate more diverse lip motions. In order to save the lip feature in the lip-value memory, we employ reconstruction loss between the recalled lip feature $\hat{f}_{{lip;valAdr}}$ and the lip feature given as a query ${f}_{ {lip}}$ as follows: 
\begin{equation}
\mathcal{L}_{store}=\|{f}_{ {lip}}-\hat{f}_{ {lip;valAdr}}\|_{2}^{2} .
\end{equation}
Through $\mathcal{L}_{store}$, the model learns to embed representative lip features of 5 sequential ground truth frames in the slots attended by value addresses.

\subsubsection{Aligning key address with value address}
After storing the lip features in the lip-value memory, we should be able to retrieve the corresponding lip feature when an audio feature is given as a query. This is how the memory network works at inference time when there are no matching ground truth images to extract lip features from. 
We obtain key address in the same way as the value address, replacing lip feature $f_{lip}$ with the audio feature $f_{aud}$ and lip-value memory $\mathbf M_{lip}$ with audio-key memory $\mathbf M_{aud}$ as follows:  
\begin{equation}
d_{aud}^{i}=\frac{m_{aud}^{i} \cdot f_{aud}}{\|m_{aud}^{i}\|_{2} \cdot\|f_{aud}\|_{2}} ,
\end{equation}
\begin{equation}
\alpha_{aud}^{i}=\frac{\exp (\kappa \cdot d_{aud}^{i})}{\sum_{j=1}^{S} \exp (\kappa \cdot d_{aud}^{j})} ,
\end{equation}
\begin{equation}
A_{aud}=\{\alpha_{aud}^{1}, \alpha_{aud}^{2}, \ldots \alpha_{aud}^{S}\} .
\end{equation}
 We align key address with value address through key-value address alignment loss:    
\begin{equation}
\mathcal{L}_{ {align}}=D_{K L}(A_{lip} \| A_{aud} ) ,
\end{equation}
which is KL divergence between the two address vectors. By aligning key address and value address obtained from audio and video pairs that are in sync, both of them point to equivalent slots in the lip-value memory.  
Therefore, we can obtain lip features by using key addresses to retrieve information saved in the lip-value memory: 
\begin{equation}
\hat{f}_{ {lip;keyAdr}}=A_{ {aud}} \cdot \mathbf{M}_{ {lip}} ,
\end{equation} 
where $\hat{f}_{ {lip;keyAdr}} \in \mathbb{R}^{C}$ is the recalled lip feature from key address.
It contains lip movement related features corresponding to the input audio, acting as a strong bridge between audio and video in synthesizing the mouth region. As the decoder can take advantage of the additional visual hints on the lip movements, both visual quality and lip-sync quality can be enhanced. Also, learning audio-visual alignment earlier in the generation step imposes a stronger lip synchronization.

\subsection{Video Synthesis}
 Identity encoder extracts identity feature $f_{I}$ from a random reference frame concatenated with a pose-prior (target face with lower-half masked) along the channel axis. The pose-prior is crucial as it guides the model to generate the lower half mouth region that fits the upper half pose, reducing artifacts when pasting back to the original video \cite{kr2019towards}. The recalled lip feature from the key address is channel-wise concatenated with the audio feature and injected into the decoder $G$. The decoder has a U-Net-like architecture \cite{ronneberger2015u} with multi-scale intermediate features concatenated with those from the identity encoder, one after every up-sampling operation. This skip-connection is to ensure that the input identity and pose features are preserved. 

 At the inference time, we take recalled lip features from key addresses as shown in Fig.\ref{fig:1} (b). At training, we additionally use lip features extracted directly from the lip encoder as shown in Fig.\ref{fig:1} (a),  
\begin{equation} 
{\hat{I}}_{{g}}=G(\hat{f}_{lip;keyAdr} \oplus f_{aud}, f_{I}) , 
\end{equation}
\begin{equation} 
\hat{I}_{{G}}=G({f}_{lip} \oplus f_{aud}, f_{I}) , 
\end{equation}
where ${\hat{I}}_{g}$ is a frame generated with a recalled lip feature from a key address and $\hat{I}_{G}$ is generated with a lip feature directly from the lip encoder. Although only ${\hat{I}}_{g}$ is used at inference, we additionally adopt $\hat{I}_{G}$ during training in loss computation so that the lip encoder learns to extract meaningful features related to the lip movement from the face sequence. 

We design our generation loss functions to increase visual quality and lip-sync quality. Reconstruction loss and perceptual loss are pertinent to visual quality, and audio-visual sync loss and visual-visual sync loss are related to lip-sync quality. Note that we compute generation loss with regards to both ${\hat{I}}_{g}$ and ${\hat{I}}_{G}$.

\begin{table*}[]
\renewcommand{\arraystretch}{1.3}
\renewcommand{\tabcolsep}{3.5mm}
\resizebox{0.9999\linewidth}{!}{
\begin{tabular}{ccccccccccc} 
\Xhline{3\arrayrulewidth}
  & \multicolumn{5}{c}{LRW} & \multicolumn{5}{c}{LRS2} \\ \cmidrule(lr){2-6} \cmidrule(lr){7-11}
 Method &  PSNR & SSIM & LMD & LSE-D & LSE-C \,\, & \,\, PSNR & SSIM & LMD & LSE-D & LSE-C \\ \hline
 \rule{0pt}{11.0pt} ATVGnet & 31.409 & 0.781 & 1.894 & 7.664 & 5.735 \,\, & \,\, 30.427 & 0.735 & 2.549 & 8.223 & 5.584  \\
 3D Identity Mem & 30.725 & 0.745 & 1.659 & 8.991 & 3.963 \,\, & \,\, 29.867 & 0.696 & 2.170 & 9.263 & 4.182  \\
 Wav2Lip & 32.147 & 0.875 & 1.371 & \bf 6.617 & \bf 7.237  \,\, & \,\, 31.274 & 0.837 & 1.940 & \bf 5.995 & \bf 8.797 \\
 PC-AVS & 30.440 & 0.778 & 1.462 & 7.344 & 6.420 \,\, & \,\, 29.887 & 0.747 & 1.963 & 7.301 & 6.728 \\
 Ground Truth & N/A & 1.000 & 0.000 & 6.968 & 6.876 \,\, & \,\, N/A & 1.000 & 0.000 & 6.259 & 8.247 \\\cdashline{1-11}
 \rule{0pt}{11.0pt} \bf Ours ($\mathcal{L}_{a\text{-}v}$) & 33.099 & 0.886 & 1.276 & 7.375 & 6.162 \,\, & \,\, \bf 32.681 & 0.875 & 1.440 & 6.392 & 7.835 \\
 \bf Ours ($\mathcal{L}_{v\text{-}v}$) & 33.112 & \bf 0.893 & 1.262 & 7.394 & 6.131 \,\, & \,\, 32.611 & 0.875 & 1.433 & 6.787 & 7.363 \\
 \rule{0pt}{11.0pt} \bf Ours ($\mathcal{L}_{a\text{-}v}+\mathcal{L}_{v\text{-}v}$) & \bf 33.126 & \bf 0.893 & \bf 1.253 & 7.013 & 6.619 \,\, & \,\,  32.529 & \bf 0.876 & \bf 1.387 & 6.352 & 7.925 \\
\Xhline{3\arrayrulewidth}
\end{tabular}}
\caption{Quantitative results on LRW and LRS2 test sets. The best scores in each metric are highlighted in bold. }
\vspace{-0.2cm}
\label{table:1}
\end{table*}

\subsubsection{Reconstruction Loss}
The network is trained to minimize L1 reconstruction loss between the generated frames and ground truth frames $I$ as follows: 
\begin{equation}
\mathcal{L}_{recon}=\frac{1}{N} \sum_{i=1}^{N}(\|\hat{I}^{i}_{g}-I^{i}\|_{1}+\|\hat{I}^{i}_{G}-I^{i}\|_{1}) .
 \end{equation}

\subsubsection{Generative Adversarial Loss}
We employ GAN loss \cite{goodfellow2014generative} to evaluate image realism. L1 reconstruction alone can yield blurry images or slight artifacts as it is a pixel-level loss. 
\begin{align}
    \mathcal{L}_{ {gan }} &= {\mathbb{E}}_{\hat{I} \in [\mathbf{\hat{I}}_{G}, \mathbf{\hat{I}}_{g}]}[\log (1-D(\hat{I}))] , 
\end{align}
\begin{align}
    \mathcal{L}_{ {disc
    }} &={\mathbb{E}}_{I}[\log (1-D({I}))]+{\mathbb{E}}_{\hat{I} \in [\mathbf{\hat{I}}_{G}, \mathbf{\hat{I}}_{g}]}[\log D(\hat{I})] .
\end{align}
$D$ is a quality discriminator trained on $\mathcal{L}_{ {disc}}$, penalizing on unrealistic face generation. We adopt its architecture from \cite{prajwal2020lip}. $\hat{I}$ is an image from a set of generated images with Eq. 10 and 11.

\subsubsection{Audio-Visual Sync Loss}
We use the audio-visual sync module proposed in \cite{prajwal2020lip, chung2016out}. We train the audio-visual sync module, $\mathcal{F}_a$ and $\mathcal{F}_v$, separately and do not fine-tune further on the generated frames so that it learns from clean pairs of audio and video segments. It takes a sequence of 5 generated frames (lower half only) and an audio segment $a$ corresponding to the frame sequence. It outputs audio feature $f_a$ and video feature  $f_v$ from which binary cross-entropy of cosine similarity is computed as follows: 
\begin{equation}
d_{\mathrm{sync}}(f_a,f_v)=\frac{f_a \cdot f_v} {\|f_a\|_{2} \cdot\|f_v\|_{2}} ,
\end{equation}
\begin{align}
\mathcal{L}_{\mathrm{a\text{-}v}}=-\frac{1}{N} \sum_{i=1}^{N}&(\log d_{\mathrm{sync}}(\mathcal{F}_a(a_{i}), \mathcal{F}_v(\mathbf{\hat{I}}^{\mathbf{i}}_{\mathbf{g}})) \\ &+\log d_{\mathrm{sync}}(\mathcal{F}_a(a_{i}), \mathcal{F}_v(\mathbf{\hat{I}}^{\mathbf{i}}_{\mathbf{G}}))) ,
\end{align}
where $\mathbf{\hat{I}}^{\mathbf{i}}_{\mathbf{g}}=\{\hat{I}^{n}_{g}\}_{n=i-2}^{i+2}$, and $\mathbf{\hat{I}}^{\mathbf{i}}_{\mathbf{G}}=\{\hat{I}^{n}_{G}\}_{n=i-2}^{i+2}$.  

\subsubsection{Visual-Visual Sync Loss}
We present visual-visual sync loss that can complement audio-visual sync loss by encouraging coherence in visual domain. Lip features from a sequence of generated frames and ground truth frames can be obtained from a lip encoder $E_{lip}$. As the lip encoder is trained to extract lip motion related features that can be aligned with audio features through $\mathcal{L}_{store}$ and $\mathcal{L}_{align}$, we can expect the lip encoder to act as a strong visual-visual sync module. We define visual-visual sync loss as L1 distance between the two features as follows:
\begin{align}
\mathcal{L}_{ {v\text{-}v}}&= \frac{1}{N} \sum_{i=1}^{N} (\|E_{lip}(\mathbf{\hat{I}}^{\mathbf{i}}_{g}) - E_{lip}(\mathbf{{I}}^{\mathbf{i}})\|_{1} \\&+ 
\|E_{lip}(\mathbf{\hat{I}}^{\mathbf{i}}_{G}) - E_{lip}(\mathbf{{I}}^{\mathbf{i}})\|_{1}) .
\end{align}
We freeze the lip encoder to exclude the loss, $\mathcal{L}_{v\text{-}v}$, from training the lip encoder. By further aligning generated frames with ground truth frames, sophisticated synchronization in pixel level can be achieved. 

\subsubsection{Total Loss}
The final objective is as follow: 
\begin{align}
\mathcal{L} &= \lambda_{1}\mathcal{L}_{recon} + \lambda_{2}\mathcal{L}_{a\text{-}v} + \lambda_{3}\mathcal{L}_{v\text{-}v} \\&+ \lambda_{4}\mathcal{L}_{gan} + \lambda_{5}\mathcal{L}_{store} + \lambda_{6}\mathcal{L}_{align} , 
\end{align}
where $\lambda_{n}$ is hyper-parameter weight.

\begin{figure*}[t!]
	\begin{minipage}[b]{1.0\linewidth}
		\centering
		\centerline{\includegraphics[width=14.9cm]{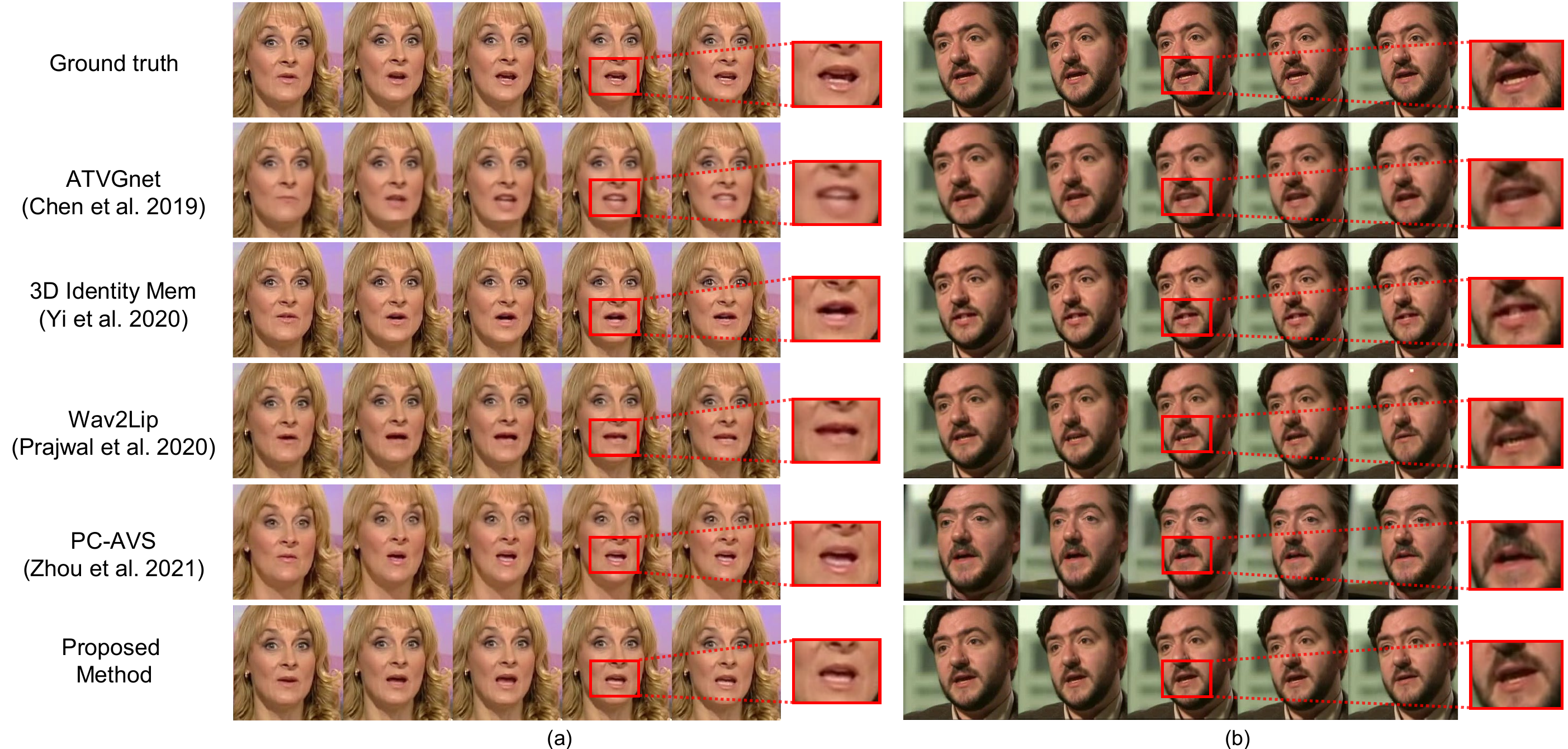}}
	\end{minipage}
	\vspace{-0.6cm}
	\caption{Comparison with state-of-the-art methods for talking face generation. Focusing on the red boxed regions, our method generates mouth that best aligns with the ground truth.  }
	\vspace{-0.3cm}
	\label{fig:2}
\end{figure*}

\section{Experiment}
\subsection{Experimental Settings}
\subsubsection{Dataset}
We train and evaluate on LRW \cite{chung2016lip} and LRS2 \cite{afouras2018deep} datasets. LRW is a word-level dataset with over 1000 utterances of 500 words. LRS2 is a sentence-level dataset with over 140,000 utterances. Both are from BBC News in the wild.

\subsubsection{Metrics}
We evaluate results using PSNR, SSIM, LMD, LSE-D, and LSE-C. PSNR and SSIM  measure visual quality and LMD, LSE-D, and LSE-C measure lip-sync quality. LMD is the distance between lip landmarks (detected using dlib \cite{king2009dlib}) of ground truth frames and those of generated frames. LSE-C and LSE-D proposed by \cite{prajwal2020lip} are confidence score (higher the better) and distance score (lower the better) between audio and video features from SyncNet \cite{chung2016out}, respectively. LSE-C and LSE-D measure correspondence between audio and visual features while LMD directly measures visual to visual coherence. For a fair comparison, we evaluate the cropped region of the face based on the face detector used in ATVGnet \cite{chen2019hierarchical}.

\subsubsection{Comparison Methods}
We compare our work with 4 state-of-the-art methods on talking face generation: ATVGnet \cite{chen2019hierarchical}, Wav2Lip \cite{prajwal2020lip}, PC-AVS \cite{zhou2021pose} and 3D Identity Mem \cite{yi2020audio}. ATVGnet generates frames conditioned on landmarks with an attention mechanism. Wav2Lip, utilized as a baseline, is a reconstruction-based method. PC-AVS employs modularized audio-visual representations of identity, pose, and speech content. 3D Identity Mem is a 3D model based method augmented with identity memory. We use open-source codes to train on the target dataset.

\subsubsection{Implementation Details}
We process video frames to face-centered crops of size 128$\times$128 at 25 fps and audio to mel-spectrogram of size 16$\times$80. Mel-spectrograms are constructed from 16kHz audio, window size 800, and hop size 200. At the inference, we use the first frame as a reference frame and the upper half of the target frame as a pose-prior. Hyper-parameters are empirically set: $\lambda_{1}$ to 10, $\lambda_{2}$, $\lambda_{3}$, $\lambda_{4}$, $\lambda_{5}$, $\lambda_{6}$ all to 0.01, and $\kappa$ to 16. We take Wav2Lip as a baseline model and add Audio-Lip Memory and lip encoder which consists of a 3D convolutional layer followed by 2D convolutional layers to encode lip motion feature.  We empirically find the optimum slot size to be 96. We first pre-train SyncNet on the target dataset and then train the framework with total loss $\mathcal{L}$ with the Adam optimizer using PyTorch. The learning rate is set to $1\times10^{-4}$, except for the discriminator, whose is $5\times10^{-4}$. We train on 8 RTX 3090 GPUs and Intel Xeon Gold CPU. 

\begin{table}[]
\renewcommand{\arraystretch}{1.3}
\renewcommand{\tabcolsep}{1.5mm}
\resizebox{0.9999\linewidth}{!}{
\begin{tabular}{cccc} 
\Xhline{3\arrayrulewidth}
 Method & Visual Quality & Lip-Sync Quality & Realness \\ \hline
 \rule{0pt}{11.0pt}  Ground Truth & 4.713 $\pm$ 0.091 & 4.871 $\pm$ 0.052 & 4.876 $\pm$ 0.041 \\\cdashline{1-4}
 ATVGnet & 2.059 $\pm$ 0.284 & 2.515 $\pm$ 0.448 & 1.803 $\pm$ 0.473 \\
 3D Identity Mem & 2.132 $\pm$ 0.399 & 1.829 $\pm$ 0.490 & 1.400 $\pm$ 0.505 \\
 Wav2Lip & 3.239 $\pm$ 0.446 & 3.929 $\pm$ 0.506 & 3.679 $\pm$ 0.592 \\
 PC-AVS & 3.108 $\pm$ 0.444 & 3.471 $\pm$ 0.491 & 3.095  $\pm$ 0.541\\ \cdashline{1-4}
\rule{0pt}{11.0pt} \bf Ours & \bf 3.582 $\pm$ \bf 0.338 & \bf 4.226 $\pm$ \bf 0.401 & \bf 3.934 $\pm$ \bf 0.480 \\
\Xhline{3\arrayrulewidth}
\end{tabular}}
\caption{Human evaluation by mean opinion scores with 95\% confidence interval on visual quality, lip-sync quality, and video realness.}
\label{table:2}
\vspace{-0.5cm}
\end{table}

\subsection{Experimental Results}

\begin{figure}[t!]
	\begin{minipage}[b]{1.0\linewidth}
		\centering
		\centerline{\includegraphics[width=7.8cm]{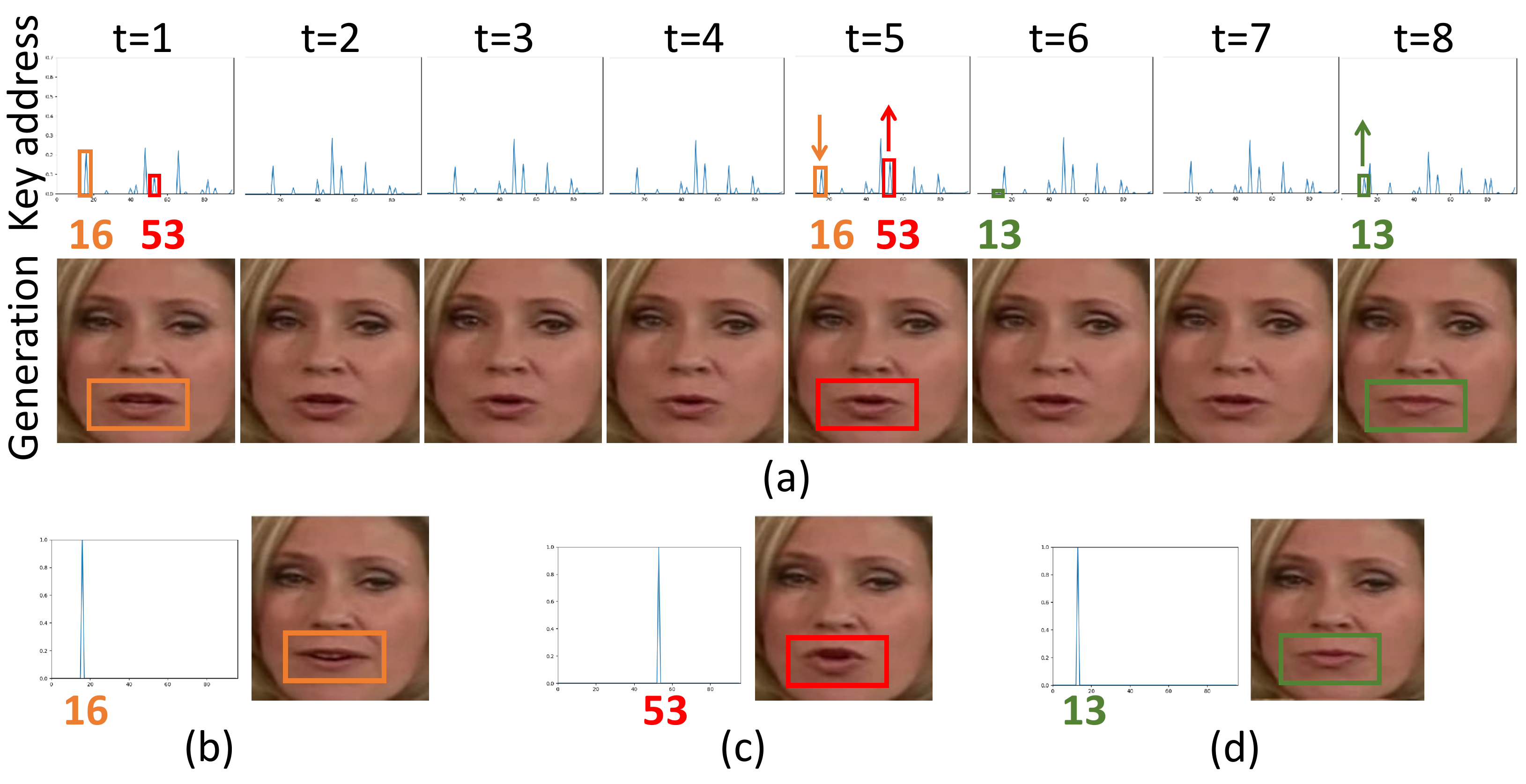}}
	\end{minipage}
	\vspace{-0.4cm}
	\caption{(a) Key addresses from audio input and corresponding generated frames in a sequence. (b), (c), and (d) Generated frames using slots 16, 53, and 13 respectively that noticeably changed its values in (a). }
	\label{fig:3}
	\vspace{-0.5cm}
\end{figure}

\subsubsection{Quantitative Results}
Table \ref{table:1} shows the quantitative comparison between other methods on LRW and LRS2 datasets. Our model generates faces with the highest PSNR, SSIM, and LMD on both datasets. Wav2Lip performs better on LSE-D and LSE-C metrics and even outperforms those of ground truth. However, as noted in \cite{zhou2021pose}, it only proves that their lip-sync results are nearly comparable to the ground truth, not better. Our LSE-D and LSE-C scores are indeed closer to the ground truth scores and we perform better on the LMD metric which is another sync metric that measures correspondence in the visual domain.
To quantify the effect of our visual-visual sync loss, we have conducted experiments using different combinations of the sync loss. As shown in Table \ref{table:1}, $\mathcal{L}_{a\text{-}v}$ has better LSE-D and LSE-C than $\mathcal{L}_{v\text{-}v}$ while $\mathcal{L}_{v\text{-}v}$ is better on PSNR, SSIM, and LMD in overall on both datasets. Such result makes sense as $\mathcal{L}_{a\text{-}v}$ is relevant to audio-visual synchronization that LSE-D and LSE-C measure while $\mathcal{L}_{v\text{-}v}$ indicates visual-visual synchronization that PSNR, SSIM, and LMD measure. It is more important to note that the two sync losses combined together yield the best performance overall on both datasets as they have complementary effects aligning different pairs of domains. Regardless of which sync loss was used, applying the memory always outperforms other methods on PSNR, SSIM, and LMD, because the memory explicitly provides visual information of the lip motion to the decoder to take advantage of. 

\begin{figure}[t!]
	\begin{minipage}[b]{1.0\linewidth}
		\centering
		\centerline{\includegraphics[width=7.4cm]{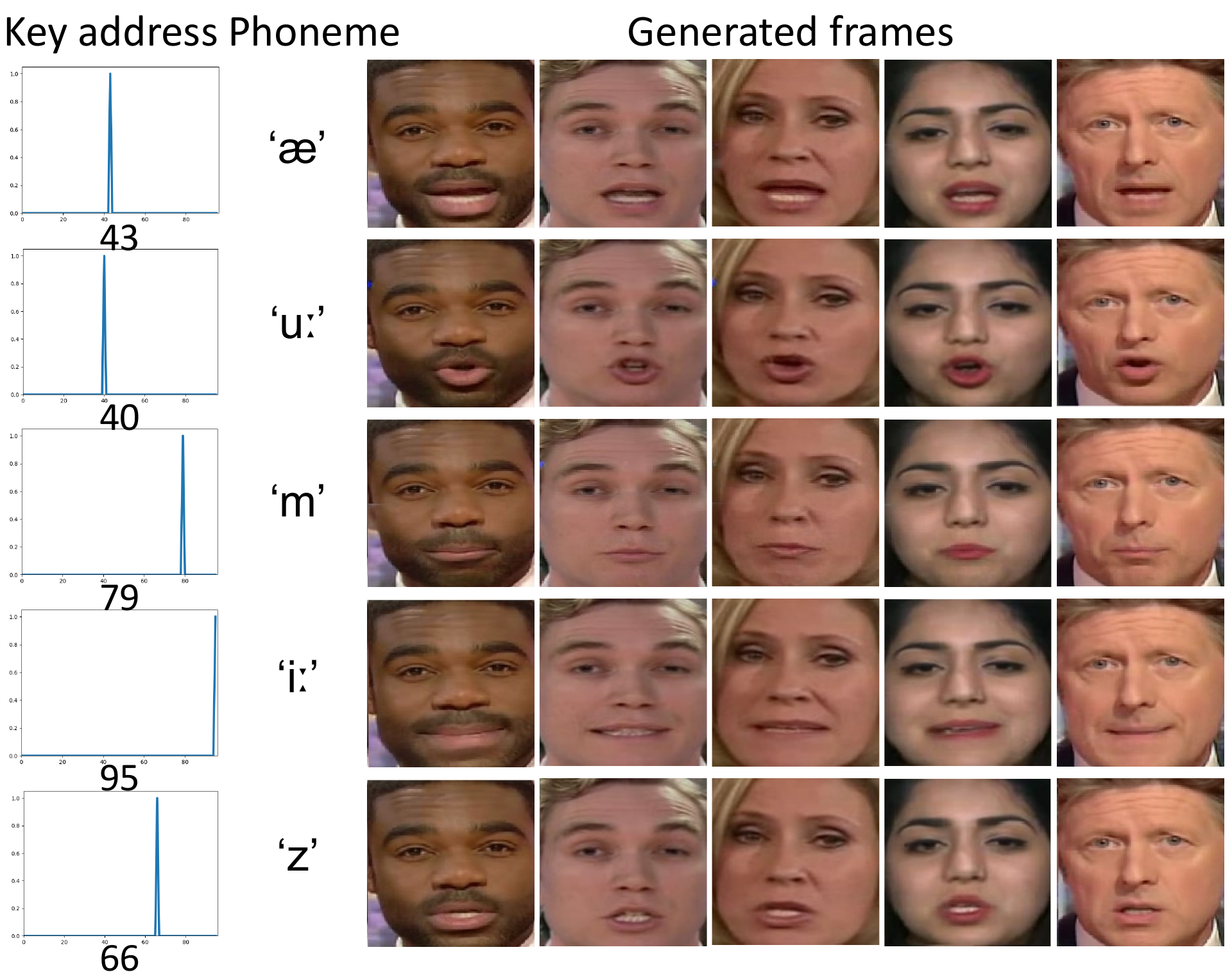}}
	\end{minipage}
	\vspace{-0.5cm}
	\caption{Generated frames using a single slot of Lip-Value Memory. Each slot contains a unique lip feature that can be associated with a phoneme.}
	\vspace{-0.2cm}
	\label{fig:4}
\end{figure}

\begin{figure}[t!]
	\begin{minipage}[b]{1.0\linewidth}
		\centering
		\centerline{\includegraphics[width=7.5cm]{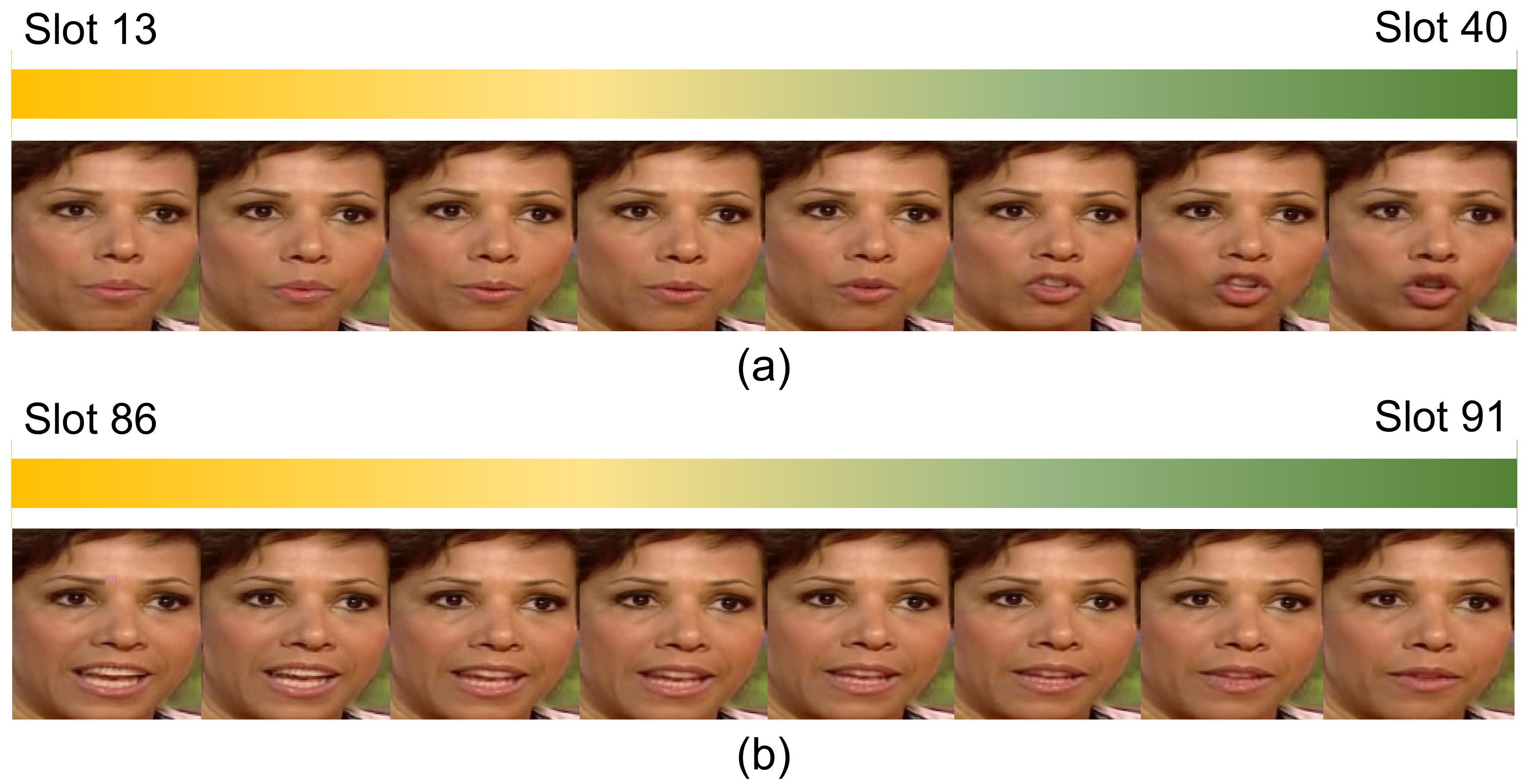}}
	\end{minipage}
	\vspace{-0.6cm}
	\caption{Interpolation of frames generated between two different slots of Lip-Value Memory. It shows that the memory address can specifically manipulate only mouth region in fine-grained level.}
	\label{fig:5}
	\vspace{-0.6cm}
\end{figure}

\subsubsection{Qualitative Results}
We compare our generation results against previous state-of-the-art methods in Fig.\ref{fig:2}. It shows that our method generates the highest quality video with mouth shapes that best match the ground truth. As ATVGnet and Identity Mem produce given one identity reference, there are restrictions to pose and expression variance so naturalness is seemingly low. PC-AVS fails to preserve the identity features of the target frame. Wav2Lip produces mouth shapes that do not exactly align with the ground truth, and there exist some artifacts. On the other hand, our method accurately captures the mouth shape including the teeth with high visual quality, as demonstrated by PSNR, SSIM, and LMD scores in Table \ref{table:1}. Such results can be contributed to the memory network allowing sophisticated and subtle lip generation and the two complementary sync losses aligning on audio and visual domains.

\subsubsection{User study}
We conduct a user study to compare generation results. 20 videos are generated using each method, 10 from the LRW test set and 10 from the LRS2 test set. 20 participants were asked to rate generated videos including ground truths to evaluate visual quality, lip-sync quality, and realness in the range of 1 to 5. As shown in Table \ref{table:2}, the scores are consistent with the quantitative results. Our method outperforms all other methods on all three criteria, especially the lip-sync quality scores. Especially the lip sync quality scores high, demonstrating effectiveness of the audio-lip memory and visual-visual synchronization loss in improving temporal coherence.

\begin{table}[t!]
    \renewcommand{\arraystretch}{1.3}
    \renewcommand{\tabcolsep}{3.5mm}
\centering
\resizebox{0.999 \linewidth}{!}{
\begin{tabular}{cccccc}
\Xhline{3\arrayrulewidth}
Slots &  PSNR & SSIM & LMD & LSE-D & LSE-C \\ \hline
24 & 32.522 & 0.873 & 1.458 & 6.442 &  7.831 \\
48 & 32.373 & 0.873 & 1.431 & 6.379 & 7.838 \\
\textbf{96} & 32.529 & \textbf{0.876} & \textbf{1.387} & \textbf{6.352} & \textbf{7.925} \\
120 & \textbf{32.655} & 0.873 & 1.469 & 6.475 & 7.785 \\ 
\Xhline{3\arrayrulewidth}
\end{tabular}}
\vspace{-0.1cm}
\caption{Ablation study on the number of slots}
\vspace{-0.6cm}
\label{table:22}
\end{table}

\subsection{Memory Analysis} 
We analyze elements stored in each slot in lip-value memory. Fig.\ref{fig:3} shows key addresses and corresponding generated frames in a sequence. The key address is generated from an audio segment pertaining to the word 'North'. We can see that the address smoothly varies as lips move. Focusing on the slots that noticeably change their address value, from t=1 to t=5, the address on the 16th slot decreases from 0.218 to 0.164 while the 53rd slot address increases from 0.097 to 0.186. To visualize the lip feature stored at each slot, we generate with \textit{silent audio} and a single slot addressed to the max as shown in Fig.\ref{fig:3} (b), (c), and (d).  We can see that a frame from slot 16 has lips drawn to the sides similar to the lip shape in t=1 and a frame from slot 53 has pursed lips as in t=5. Also, a frame from slot 13 has closed lips as in t=8 frame when the address on slot 13 suddenly increased. This result indicates that the memory well decouples lip features associated with speech sound and bestows memory with explicit control over the lip movement while keeping all other factors such as identity and pose unchanged.

We further generate frames using a single slot in Fig.\ref{fig:4}. It is possible to assign each slot with a phoneme. For example, slot 43 closely aligns with 'æ', slot 40 'u:', slot 79 'm', slot 95 'i:' and slot 66 'z'. It demonstrates that each slot contains a unique lip feature at the phoneme level and that by taking combinations of the lip features in each slot through address, diverse lip movements can be generated. 

We verify that lip shape can be smoothly interpolated between addresses on two different slots. As shown in Fig.\ref{fig:5}, as the ratio of the address varies from 13 to 40 in (a) and from 86 to 91 in (b), the lips change accordingly. Since the generation is very sensitive to the address value concerning the input audio, our method can generate subtle lip movements. 

Lastly, we perform ablation study on using a different number of slots as shown in Table \ref{table:22}. The performance gradually increases from using 24 slots to 96 slots but decreases when the slot size is further increased to 120. This indicates that a large number of slots to hold many lip features does not linearly increase the performance because it may complicate the model in aligning key and value addresses. Thus, we empirically set the optimum slot size to 96.

\section{Conclusion}
Our proposed Audio-Lip Memory extracts the lip motion related features to bridge from audio to video generation. The lip synchronization is achieved during the memory learning that aligns the audio and visual lip features, and it is further enforced by the audio-visual and the visual-visual synchronization losses.
We have verified that the lip features are stored in each memory slot at the phoneme level, and different combinations of the slots through memory addressing can yield diverse and subtle lip motions. 
Therefore, our work effectively exploits visual information of the mouth region to simultaneously achieve high visual quality and lip synchronization for talking face generation.

\section{Acknowledgements}
This work was partially supported by Genesis Lab under a research project (G01210312).

\bibliography{aaai22.bib} 

\begin{thebibliography}{39}
\providecommand{\natexlab}[1]{#1}

\bibitem[{Afouras et~al.(2018)Afouras, Chung, Senior, Vinyals, and
  Zisserman}]{afouras2018deep}
Afouras, T.; Chung, J.~S.; Senior, A.; Vinyals, O.; and Zisserman, A. 2018.
\newblock Deep audio-visual speech recognition.
\newblock \emph{IEEE transactions on pattern analysis and machine
  intelligence}.

\bibitem[{Cai et~al.(2018)Cai, Pan, Yao, Yan, and Mei}]{cai2018memory}
Cai, Q.; Pan, Y.; Yao, T.; Yan, C.; and Mei, T. 2018.
\newblock Memory matching networks for one-shot image recognition.
\newblock In \emph{Proceedings of the IEEE conference on computer vision and
  pattern recognition}, 4080--4088.

\bibitem[{Chen et~al.(2018)Chen, Li, Maddox, Duan, and Xu}]{chen2018lip}
Chen, L.; Li, Z.; Maddox, R.~K.; Duan, Z.; and Xu, C. 2018.
\newblock Lip movements generation at a glance.
\newblock In \emph{Proceedings of the European Conference on Computer Vision
  (ECCV)}, 520--535.

\bibitem[{Chen et~al.(2019)Chen, Maddox, Duan, and Xu}]{chen2019hierarchical}
Chen, L.; Maddox, R.~K.; Duan, Z.; and Xu, C. 2019.
\newblock Hierarchical cross-modal talking face generation with dynamic
  pixel-wise loss.
\newblock In \emph{Proceedings of the IEEE/CVF Conference on Computer Vision
  and Pattern Recognition}, 7832--7841.

\bibitem[{Chung and Zisserman(2016{\natexlab{a}})}]{chung2016lip}
Chung, J.~S.; and Zisserman, A. 2016{\natexlab{a}}.
\newblock Lip reading in the wild.
\newblock In \emph{Asian conference on computer vision}, 87--103. Springer.

\bibitem[{Chung and Zisserman(2016{\natexlab{b}})}]{chung2016out}
Chung, J.~S.; and Zisserman, A. 2016{\natexlab{b}}.
\newblock Out of time: automated lip sync in the wild.
\newblock In \emph{Asian conference on computer vision}, 251--263. Springer.

\bibitem[{Chung, Chung, and Kang(2019)}]{chung2019perfect}
Chung, S.-W.; Chung, J.~S.; and Kang, H.-G. 2019.
\newblock Perfect match: Improved cross-modal embeddings for audio-visual
  synchronisation.
\newblock In \emph{ICASSP 2019-2019 IEEE International Conference on Acoustics,
  Speech and Signal Processing (ICASSP)}, 3965--3969. IEEE.

\bibitem[{Das et~al.(2020)Das, Biswas, Sinha, and Bhowmick}]{das2020speech}
Das, D.; Biswas, S.; Sinha, S.; and Bhowmick, B. 2020.
\newblock Speech-driven facial animation using cascaded gans for learning of
  motion and texture.
\newblock In \emph{European Conference on Computer Vision}, 408--424. Springer.

\bibitem[{Gao and Grauman(2021)}]{gao2021visualvoice}
Gao, R.; and Grauman, K. 2021.
\newblock Visualvoice: Audio-visual speech separation with cross-modal
  consistency.
\newblock \emph{arXiv preprint arXiv:2101.03149}.

\bibitem[{Goodfellow et~al.(2014)Goodfellow, Pouget-Abadie, Mirza, Xu,
  Warde-Farley, Ozair, Courville, and Bengio}]{goodfellow2014generative}
Goodfellow, I.; Pouget-Abadie, J.; Mirza, M.; Xu, B.; Warde-Farley, D.; Ozair,
  S.; Courville, A.; and Bengio, Y. 2014.
\newblock Generative adversarial nets.
\newblock \emph{Advances in neural information processing systems}, 27.

\bibitem[{Jamaludin, Chung, and Zisserman(2019)}]{jamaludin2019you}
Jamaludin, A.; Chung, J.~S.; and Zisserman, A. 2019.
\newblock You said that?: Synthesising talking faces from audio.
\newblock \emph{International Journal of Computer Vision}, 127(11): 1767--1779.

\bibitem[{Kaiser et~al.(2017)Kaiser, Nachum, Roy, and
  Bengio}]{kaiser2017learning}
Kaiser, {\L}.; Nachum, O.; Roy, A.; and Bengio, S. 2017.
\newblock Learning to remember rare events.
\newblock \emph{arXiv preprint arXiv:1703.03129}.

\bibitem[{Kim, Park, and Ro(2021)}]{kim2021robust}
Kim, J.~U.; Park, S.; and Ro, Y.~M. 2021.
\newblock Robust Small-Scale Pedestrian Detection With Cued Recall via Memory
  Learning.
\newblock In \emph{Proceedings of the IEEE/CVF International Conference on
  Computer Vision}, 3050--3059.

\bibitem[{Kim et~al.(2021)Kim, Hong, Park, and Ro}]{kim2021multi}
Kim, M.; Hong, J.; Park, S.~J.; and Ro, Y.~M. 2021.
\newblock Multi-modality associative bridging through memory: Speech sound
  recollected from face video.
\newblock In \emph{Proceedings of the IEEE/CVF International Conference on
  Computer Vision}, 296--306.

\bibitem[{Kim and Ro(2021)}]{kim2021m}
Kim, S.; and Ro, Y.~M. 2021.
\newblock M-CAM: Visual Explanation of Challenging Conditioned Dataset with
  Bias-reducing Memory.
\newblock In \emph{The 32nd British Machine Vision Conference, BMVC 2021}.
  British Machine Vision Association (BMVA).

\bibitem[{King(2009)}]{king2009dlib}
King, D.~E. 2009.
\newblock Dlib-ml: A machine learning toolkit.
\newblock \emph{The Journal of Machine Learning Research}, 10: 1755--1758.

\bibitem[{KR et~al.(2019)KR, Mukhopadhyay, Philip, Jha, Namboodiri, and
  Jawahar}]{kr2019towards}
KR, P.; Mukhopadhyay, R.; Philip, J.; Jha, A.; Namboodiri, V.; and Jawahar, C.
  2019.
\newblock Towards automatic face-to-face translation.
\newblock In \emph{Proceedings of the 27th ACM International Conference on
  Multimedia}, 1428--1436.

\bibitem[{Lee et~al.(2021)Lee, Kim, Choi, Kim, and Ro}]{lee2021video}
Lee, S.; Kim, H.~G.; Choi, D.~H.; Kim, H.-I.; and Ro, Y.~M. 2021.
\newblock Video Prediction Recalling Long-term Motion Context via Memory
  Alignment Learning.
\newblock In \emph{Proceedings of the IEEE/CVF Conference on Computer Vision
  and Pattern Recognition}, 3054--3063.

\bibitem[{Lee et~al.(2018)Lee, Sung, Yu, and Kim}]{lee2018memory}
Lee, S.; Sung, J.; Yu, Y.; and Kim, G. 2018.
\newblock A memory network approach for story-based temporal summarization of
  360 videos.
\newblock In \emph{Proceedings of the IEEE Conference on Computer Vision and
  Pattern Recognition}, 1410--1419.

\bibitem[{Miller et~al.(2016)Miller, Fisch, Dodge, Karimi, Bordes, and
  Weston}]{miller2016key}
Miller, A.; Fisch, A.; Dodge, J.; Karimi, A.-H.; Bordes, A.; and Weston, J.
  2016.
\newblock Key-value memory networks for directly reading documents.
\newblock \emph{arXiv preprint arXiv:1606.03126}.

\bibitem[{Mittal and Wang(2020)}]{mittal2020animating}
Mittal, G.; and Wang, B. 2020.
\newblock Animating face using disentangled audio representations.
\newblock In \emph{Proceedings of the IEEE/CVF Winter Conference on
  Applications of Computer Vision}, 3290--3298.

\bibitem[{Nagrani et~al.(2020)Nagrani, Chung, Albanie, and
  Zisserman}]{nagrani2020disentangled}
Nagrani, A.; Chung, J.~S.; Albanie, S.; and Zisserman, A. 2020.
\newblock Disentangled speech embeddings using cross-modal self-supervision.
\newblock In \emph{ICASSP 2020-2020 IEEE International Conference on Acoustics,
  Speech and Signal Processing (ICASSP)}, 6829--6833. IEEE.

\bibitem[{Pei et~al.(2019)Pei, Zhang, Wang, Ke, Shen, and Tai}]{pei2019memory}
Pei, W.; Zhang, J.; Wang, X.; Ke, L.; Shen, X.; and Tai, Y.-W. 2019.
\newblock Memory-attended recurrent network for video captioning.
\newblock In \emph{Proceedings of the IEEE/CVF Conference on Computer Vision
  and Pattern Recognition}, 8347--8356.

\bibitem[{Prajwal et~al.(2020)Prajwal, Mukhopadhyay, Namboodiri, and
  Jawahar}]{prajwal2020lip}
Prajwal, K.; Mukhopadhyay, R.; Namboodiri, V.~P.; and Jawahar, C. 2020.
\newblock A lip sync expert is all you need for speech to lip generation in the
  wild.
\newblock In \emph{Proceedings of the 28th ACM International Conference on
  Multimedia}, 484--492.

\bibitem[{Ronneberger, Fischer, and Brox(2015)}]{ronneberger2015u}
Ronneberger, O.; Fischer, P.; and Brox, T. 2015.
\newblock U-net: Convolutional networks for biomedical image segmentation.
\newblock In \emph{International Conference on Medical image computing and
  computer-assisted intervention}, 234--241. Springer.

\bibitem[{Song et~al.(2020)Song, Wu, Qian, He, and Loy}]{song2020everybody}
Song, L.; Wu, W.; Qian, C.; He, R.; and Loy, C.~C. 2020.
\newblock Everybody's talkin': Let me talk as you want.
\newblock \emph{arXiv preprint arXiv:2001.05201}.

\bibitem[{Song et~al.(2018)Song, Zhu, Li, Wang, and Qi}]{song2018talking}
Song, Y.; Zhu, J.; Li, D.; Wang, X.; and Qi, H. 2018.
\newblock Talking face generation by conditional recurrent adversarial network.
\newblock \emph{arXiv preprint arXiv:1804.04786}.

\bibitem[{Thies et~al.(2020)Thies, Elgharib, Tewari, Theobalt, and
  Nie{\ss}ner}]{thies2020neural}
Thies, J.; Elgharib, M.; Tewari, A.; Theobalt, C.; and Nie{\ss}ner, M. 2020.
\newblock Neural voice puppetry: Audio-driven facial reenactment.
\newblock In \emph{European Conference on Computer Vision}, 716--731. Springer.

\bibitem[{Vougioukas, Petridis, and Pantic(2020)}]{vougioukas2020realistic}
Vougioukas, K.; Petridis, S.; and Pantic, M. 2020.
\newblock Realistic speech-driven facial animation with gans.
\newblock \emph{International Journal of Computer Vision}, 128(5): 1398--1413.

\bibitem[{Wang, Mallya, and Liu(2021)}]{wang2021one}
Wang, T.-C.; Mallya, A.; and Liu, M.-Y. 2021.
\newblock One-shot free-view neural talking-head synthesis for video
  conferencing.
\newblock In \emph{Proceedings of the IEEE/CVF Conference on Computer Vision
  and Pattern Recognition}, 10039--10049.

\bibitem[{Weston, Chopra, and Bordes(2014)}]{weston2014memory}
Weston, J.; Chopra, S.; and Bordes, A. 2014.
\newblock Memory networks.
\newblock \emph{arXiv preprint arXiv:1410.3916}.

\bibitem[{Yi et~al.(2020)Yi, Ye, Zhang, Bao, and Liu}]{yi2020audio}
Yi, R.; Ye, Z.; Zhang, J.; Bao, H.; and Liu, Y.-J. 2020.
\newblock Audio-driven talking face video generation with learning-based
  personalized head pose.
\newblock \emph{arXiv preprint arXiv:2002.10137}.

\bibitem[{Yu et~al.(2020)Yu, Yu, Li, and Ling}]{yu2020multimodal}
Yu, L.; Yu, J.; Li, M.; and Ling, Q. 2020.
\newblock Multimodal Inputs Driven Talking Face Generation With
  Spatial--Temporal Dependency.
\newblock \emph{IEEE Transactions on Circuits and Systems for Video
  Technology}, 31(1): 203--216.

\bibitem[{Zhang et~al.(2021)Zhang, Li, Ding, and Fan}]{zhang2021flow}
Zhang, Z.; Li, L.; Ding, Y.; and Fan, C. 2021.
\newblock Flow-Guided One-Shot Talking Face Generation With a High-Resolution
  Audio-Visual Dataset.
\newblock In \emph{Proceedings of the IEEE/CVF Conference on Computer Vision
  and Pattern Recognition}, 3661--3670.

\bibitem[{Zhou et~al.(2019)Zhou, Liu, Liu, Luo, and Wang}]{zhou2019talking}
Zhou, H.; Liu, Y.; Liu, Z.; Luo, P.; and Wang, X. 2019.
\newblock Talking face generation by adversarially disentangled audio-visual
  representation.
\newblock In \emph{Proceedings of the AAAI Conference on Artificial
  Intelligence}, volume~33, 9299--9306.

\bibitem[{Zhou et~al.(2021)Zhou, Sun, Wu, Loy, Wang, and Liu}]{zhou2021pose}
Zhou, H.; Sun, Y.; Wu, W.; Loy, C.~C.; Wang, X.; and Liu, Z. 2021.
\newblock Pose-controllable talking face generation by implicitly modularized
  audio-visual representation.
\newblock In \emph{Proceedings of the IEEE/CVF Conference on Computer Vision
  and Pattern Recognition}, 4176--4186.

\bibitem[{Zhou et~al.(2020)Zhou, Han, Shechtman, Echevarria, Kalogerakis, and
  Li}]{zhou2020makelttalk}
Zhou, Y.; Han, X.; Shechtman, E.; Echevarria, J.; Kalogerakis, E.; and Li, D.
  2020.
\newblock MakeltTalk: speaker-aware talking-head animation.
\newblock \emph{ACM Transactions on Graphics (TOG)}, 39(6): 1--15.

\bibitem[{Zhu et~al.(2020)Zhu, Huang, Li, Zheng, and He}]{inproceedings}
Zhu, H.; Huang, H.; Li, Y.; Zheng, A.; and He, R. 2020.
\newblock Arbitrary Talking Face Generation via Attentional Audio-Visual
  Coherence Learning.
\newblock 2334--2340.

\bibitem[{Zhu et~al.(2019)Zhu, Pan, Chen, and Yang}]{zhu2019dm}
Zhu, M.; Pan, P.; Chen, W.; and Yang, Y. 2019.
\newblock Dm-gan: Dynamic memory generative adversarial networks for
  text-to-image synthesis.
\newblock In \emph{Proceedings of the IEEE/CVF Conference on Computer Vision
  and Pattern Recognition}, 5802--5810.

\end{thebibliography}
\end{document}